\documentclass[runningheads, envcountsame, a4paper]{llncs}

\usepackage[misc]{ifsym}
\usepackage[hyphens]{url}
\usepackage{microtype}
\usepackage[pdftex]{graphicx}
\usepackage{epstopdf}
\usepackage{amsmath,amssymb}

\newcommand{\ra}{\rightarrow}

\newcommand{\bfx}{\mathbf{x}}

\begin{document}
\title{A General Machine Learning Framework for Survival Analysis}
\titlerunning{A General Machine Learning Framework for Survival Analysis}

\author{Andreas Bender(\Letter)\inst{1}\orcidID{0000-0001-5628-8611} \and
David R\"ugamer\inst{1}\orcidID{0000-0002-8772-9202} \and
Fabian Scheipl\inst{1}\orcidID{0000-0001-8172-3603} \and Bernd Bischl\inst{1}\orcidID{0000-0001-6002-6980}}
\authorrunning{A. Bender et al.}

\toctitle{A General Machine Learning Framework for Survival Analysis}
\tocauthor{A. Bender et al.}

\institute{LMU Munich, Department of Statistics, Ludwigstr. 33, 80539 Munich\\
\email{andreas.bender@stat.uni-muenchen.de}\\}
\maketitle              
\begin{abstract}
The modeling of time-to-event data, also known as survival analysis, requires specialized methods that can deal with censoring and truncation, time-varying features and effects, and that extend to settings with multiple competing events. However, many machine learning methods for survival analysis only consider the standard setting with right-censored data and proportional hazards assumption. The methods that do provide extensions usually address at most a subset of these challenges and often require specialized software that can not be integrated into standard machine learning workflows directly. In this work, we present a very general machine learning framework for time-to-event analysis that uses a data augmentation strategy to reduce complex survival tasks to standard Poisson regression tasks. This reformulation is based on well developed statistical theory. With the proposed approach, any algorithm that can optimize a Poisson (log-)likelihood, such as gradient boosted trees, deep neural networks, model-based boosting and many more can be used in the context of time-to-event analysis. The proposed technique does not require any assumptions with respect to the distribution of event times or the functional shapes of feature and interaction effects.  Based on the proposed framework  we develop new methods that are competitive with specialized state of the art approaches in terms of accuracy, and versatility, but with comparatively small investments of programming effort or requirements for specialized methodological know-how.

\end{abstract}
\section{Introduction}
Survival analysis is a branch of statistics that provides a framework for the analysis of time-to-event data, i.e., the outcome is defined by the time it takes until an event occurs. Analysis of such data requires specialized techniques because, in contrast to standard regression or classification tasks,

\begin{itemize}
  \item[(a)] the outcome can often not be observed fully (censoring, truncation),
  \item[(b)] the features can change their value during the observation period (time-varying features (TVF),
  \item[(c)] the association of the feature(s) with the outcome changes over time (time-varying effects (TVE)),
  \item[(d)] one or more other events occur that make it impossible to observe the event of interest (competing risks (CR)),
  \item[(e)] more generally, in a multi-state setting, observation units can move from and to different states (multi-state models (MSM)).
\end{itemize}

\begin{figure}[!ht]
\includegraphics[width = 1\textwidth]{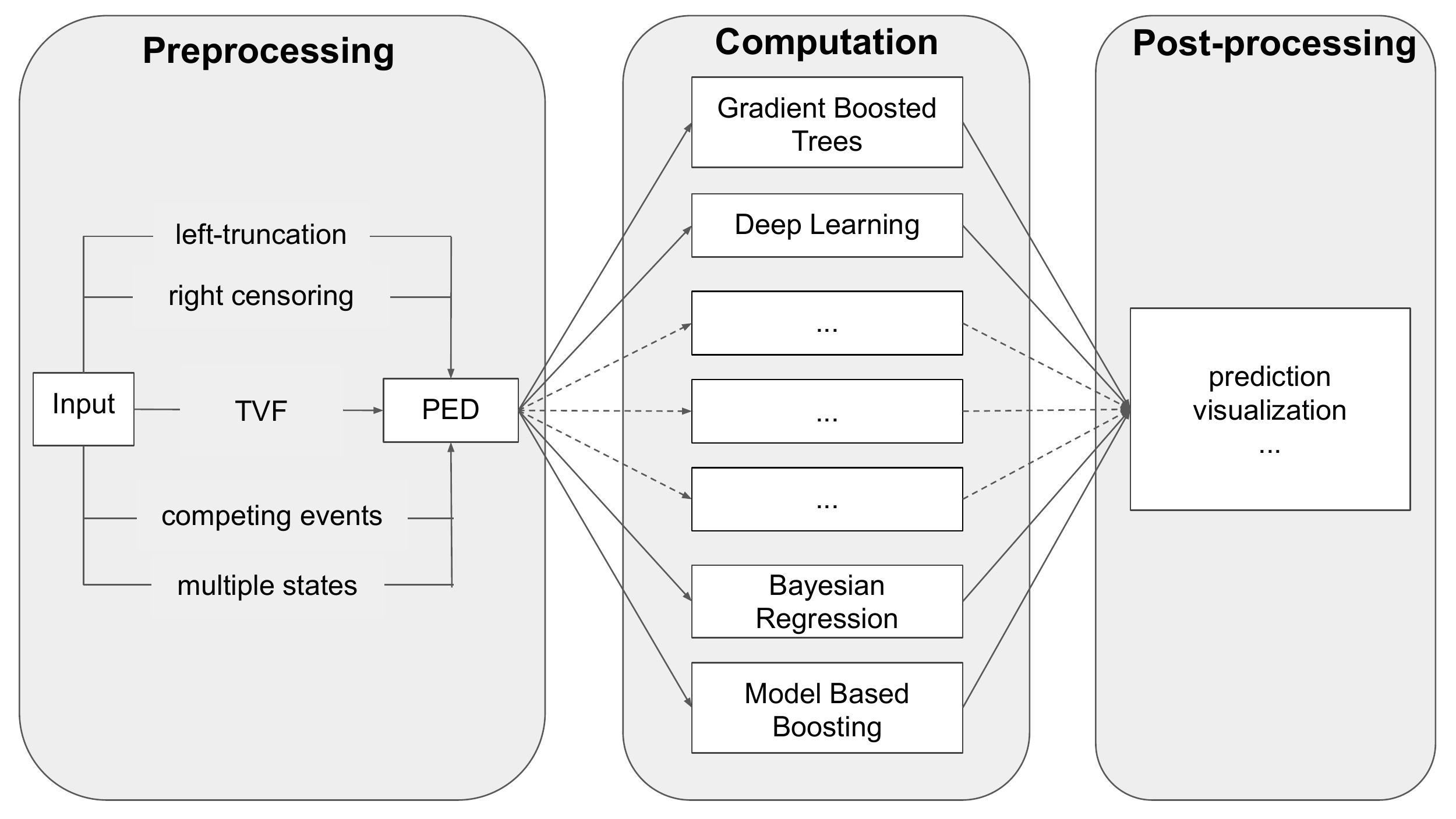}
\caption{An abstraction of survival analysis for different tasks. The structure
of the piece-wise exponential data (PED) depends on the task requirements, e.g.,
left truncation or competing risks. Given the appropriate pre-processing, the
estimation step is computationally independent of the survival task, except for an appropriate use of interaction terms.}
\label{fig:framework}
\end{figure}

Failure to take these issues into consideration usually results in biased estimates, incorrect interpretation of feature effects on the outcome, loss of predictive accuracy or a combination thereof.
In this work, we use a reformulation of the survival task to a standard regression task that provides a holistic approach to survival analysis. Within this framework, censoring, truncation and time-varying features (TVF) can be incorporated by specific data transformations and extensions to time-varying effects (TVE) as well as competing risks and multi-state models can be re-expressed in terms of interaction effects. This abstraction of the survival task away from specialized algorithms is illustrated schematically in Figure \ref{fig:framework}. Task-appropriate pre-processing (leftmost subgraph) yields a standardized data format that allows the estimation of feature-conditional hazard rates using any learning algorithm that can minimize the negative Poisson log-likelihood, such as, GBT, deep neural networks (DNN), regularization based methods, and others (middle subgraph).

\paragraph{Our contributions}
\ \\
We define a general machine learning framework for survival analysis based on piece-wise exponential models (cf. Section \ref{sec:methods}). Within this framework, different concepts specific to time-to-event data analysis can be understood in terms of data augmentation and inclusion of interaction terms. By re-expressing the survival task as a Poisson regression task, a large variety of algorithms become available for survival analysis. Based on the proposed approach, we implement a gradient boosted trees algorithm with comparatively low development effort and show that it achieves state-of-the-art performance (cf. Section \ref{sec:experiments}).

\paragraph{Related work}
\ \\
The machine learning community has developed many highly efficient methods for high-dimensional settings in different domains, including survival analysis. The individual methods and implementations, however, often only support a subset of the cases relevant for time-to-event analysis mentioned above. For example, the random survival forest (RSF) proposed in \cite{ishwaran_random_2008} was later extended to the competing risks setting \cite{ishwaran_random_2014}, but does not support left-truncation, TVF and TVE, or multistate models. Another popular implementation of random forests \cite{wright_ranger:_2017} only supports right censored data and proportional hazards models. An extension of RSF, the oblique RSF (ORSF, \cite{jaeger_oblique_2019}) was shown to outperform other RSF based algorithms, but has the same limitations. With respect to TVF and TVE, a review of tree- and forest-based methods for survival analysis stated that ``the modeling of time–varying features and time–varying effects deserves much more attention'' \cite{bou-hamad_review_2011}. Similarly, a more recent review of machine learning methods for survival analysis \cite{wang_machine_2019} only lists the time-dependent Cox model \cite[Ch. 9.2]{kleinmoeschberger_2006}, and L1- and L2-regularized extensions thereof, as a possibility for the inclusion of TVF.

Deep learning based methods for time-to-event data have also received much attention lately. An early use of neural networks for Cox type models was proposed by \cite{faraggi_neural_1995}. More recently, \cite{ranganath_deep_2016} presented a framework for deep single event survival analysis based on a joint latent process for features and survival times using deep exponential families. For competing risks data, a deep learning framework based on Gaussian processes was described in \cite{alaa_2017}. Another recent framework is DeepHit, which can handle competing risks using a custom loss function \cite{lee_deephit_2018} and was extended to handle TVF \cite{lee_dynamic-deephit_2020}, but did not discuss left-truncation, multistate models and TVE.

Boosting has also been a popular technique for high-dimensional survival analysis. For example, \cite{binder_boosting_2009} propose a Cox-type boosting approach for the estimation of proportional sub-distribution hazards. A flexible multi-state model based on the stratified Cox partial likelihood in the context of model-based boosting \cite{Hothorn2006} is presented in \cite{reulen_boosting_2015}. Furthermore, an implementation of gradient boosted trees (GBT) for the Cox PH model is also available for the popular XGBoost implementation \cite{chen_xgboost:_2016}, which was also shown to perform well compared with the ORSF \cite{jaeger_oblique_2019}. Recently, \cite{lee_boosted_2019} derived a custom algorithm for gradient boosted trees that support TVF and demonstrate that their inclusion improves predictive performance compared to boosting algorithms that don't take TVF into account.

Compared to methods based on Cox regression, few publications have developed methods based on the piece-wise exponential model, on which the framework proposed here is based. Among them is an early application of neural networks to survival analysis suggested in \cite{liestbl_survival_1994} and extended by \cite{biganzoli_general_2002}. The latter offers a general framework based on the representation of generalized linear models via feed forward neural networks, but does not discuss MSM. Piece-wise exponential trees with TVF and splits based on the piece-wise exponential survival function were suggested by \cite{huang_piecewise_1998}. A spline based estimation of the hazard function was discussed in \cite{Cai2002}, which could also be represented via neural networks (cf. \cite{fornili_piecewise_2014}). A flexible estimation of piece-wise-exponential model based multi-state models with shared effects using structured fusion Lasso was developed in \cite{sennhenn-reulen_structured_2016}. All of these methods can be viewed as special cases within the proposed framework. For example, \cite{biganzoli_general_2002} could be extended to different neural network architectures and MSMs, \cite{huang_piecewise_1998} could be extended to forests.

\section{Survival Analysis as Poisson Regression}\label{sec:methods}
In the context of survival analysis, an observation usually consists of a tuple $(t_i, \delta_i, \bfx_i)$, where $t_i$ is the observed event time for observation unit $i=1,\ldots,n$, $\delta_i \in \{0,1\}$ is the event- or status-indicator (i.e. 1 if event occurred, 0 if the observation of censored) and $\bfx_i$ is the $p$-dimensional feature vector. The presence of censoring requires special estimation techniques, as the time-to-event can not be observed when censoring occurs before the event of interest. Thus $t_i = \min(T_i, C_i)$, where $T_i$ and $C_i$ random variables of the event time and censoring time, respectively. A classic example is the time until death when censoring occurs as patients drop out of the study (unrelated to the event of interest, $T_i\perp C_i$). Left-truncation occurs when the event of interest already occurred before the subject could be included into the sample and thus presents a form of sampling bias.
In some settings, another event could preclude observation of the event of interest or change the probability of its occurrence. In this case we speak of competing risks (CR), thus the observation consists of $(t_i, \delta_i, k, \bfx_i)$, where $k = 1,\ldots, K$ indicates the type of event that occurred at, $t_i$ if $\delta_i = 1$. More generally, there might be multiple states that the observation units can transition from and to. We then speak of multi-state models (MSM) and $k$ is an indicator for different transitions (cf. Eq. \eqref{eq:cshazard}).

In general, the goal of survival analysis is to estimate the conditional distribution of event times defined by the survival probability $S(t|\bfx) = P(T > t|\bfx)$. While some methods focus on the estimation of $S(t|\bfx)$ directly, it is often more convenient to estimate the (log-)hazard

\begin{equation}\label{eq:hazard}
\lambda(t|\bfx) := \lim\limits_{\Delta t \to 0} \frac{P(t\leq T <t+\Delta t | T\geq t, \bfx)}{\Delta t}\,
\end{equation}
from which $S(t|\bfx)$ follows as
\begin{equation}
S(t|\bfx) = \exp\left(-\int_0^t \lambda(s|\bfx)\mathrm{d}s\right).
\end{equation}
Here we represent \eqref{eq:hazard} via
\begin{equation}\label{eq:phmodel}
\lambda(t|\bfx(t)) = \exp(g(\bfx(t), t)),
\end{equation}
where $g$ is a general function of potentially TVF $\bfx(t)$, that can include high-order feature interactions, non-linearity and time-dependence of feature effects (TVE) via an interaction with $t$.\\

In this work, we approximate \eqref{eq:phmodel} using the piece-wise exponential model \cite{Friedman1982}. Let $t_i$ the observed event or censoring time and $\delta_i \in \{0,1\}$ the respective censoring or event indicator for observation units $i=1,\ldots,n$. The distribution of censoring times can depend on features but is assumed to be independent of the event time process $T$. By partitioning the follow-up, i.e., the time span under investigation, into $j=1,\ldots,J$ intervals with cut-points $\kappa_0 = 0 < \cdots < \kappa_J$ and partitions
$(\kappa_0, \kappa_1],\ldots, (\kappa_{j-1},\kappa_{j}],\ldots (\kappa_{J-1},\kappa_{J}]$,
we can rewrite \eqref{eq:phmodel} using piece-wise constant hazard rates
\begin{align}\label{eq:pcmodel}
\lambda(t| \bfx_i(t))
  & \equiv \exp(g(\bfx_{ij}, t_j)):=\lambda_{ij},\ \ \forall t \in (\kappa_{j-1}, \kappa_j],
\end{align}
with $t_j$ a representation of time in interval $j$, e.g., $t_j:=\kappa_j$ and  $\bfx_{ij}$ the value of the TVF in interval $j$. Depending on the desired resolution, additional cut-points can be introduced at each time point at which feature values are updated, otherwise multiple feature values have to be aggregated in one interval. This model assumes that only the current value of $\bfx_{ij}$ affects the hazard in interval $j$, but more sophisticated approaches have been suggested within this framework that take into account the entire history of TVF \cite{Bender2018a}.
Piece-wise constant hazards imply piece-wise exponential log-likelihood contributions
\begin{equation}\label{eq:ll}
\ell_i
  = \log(\lambda(t_i;\bfx_i)^{\delta_i}S(t_i;\bfx_i))\\
  = \sum_{j=1}^{J_i}\left(\delta_{ij}\log\lambda_{ij} -  \lambda_{ij}t_{ij}\right),
\end{equation}
where $J_i$ is the last interval in which observation unit $i$ was observed, such that $t_{i} \in (\kappa_{J_i-1},\kappa_{J_i}]$ and

\begin{equation}\label{eq:ped_status}
\delta_{ij} = \begin{cases}1 & t_i \in (\kappa_{j-1}, \kappa_j] \wedge \delta_i = 1\\0 & \text{else}\end{cases},\
t_{ij} = \begin{cases}t_{i}-\kappa_{j-1} & \delta_{ij}=1\\ \kappa_{j}-\kappa_{j-1}& \text{else}\end{cases}.
\end{equation}
Concrete examples for the type of data transformations required to obtain \eqref{eq:ped_status} for right-censored data (including TVF) are provided in \cite{Bender2018b} (cf. Tables \ref{tab:data-trafo} and \ref{tab:exampleData}.

Using the working assumption  $\delta_{ij}\stackrel{iid}{\sim}Poisson(\mu_{ij}=\lambda_{ij}t_{ij})$ and with $f(\delta_{ij})$ the Poisson density function, \cite{Friedman1982} showed that the Poisson log-likelihood

\begin{equation}\label{eq:poisson-ll}
\ell_i
   = \log\left(\prod_{j=1}^{J_i} f(\delta_{ij})\right)
   = \sum_{j=1}^{J_i} ( \delta_{ij}\log \lambda_{ij} + \delta_{ij}\log t_{ij} - \lambda_{ij}t_{ij})
\end{equation}
is proportional to \eqref{eq:ll} and therefore the former can be minimized using Poisson regression. Note that \eqref{eq:poisson-ll} can be directly extended to the setting with left-truncated event times \cite{guo_event-history_1993} by replacing $j=1$ with $j_i$, the first interval in which observation unit $i$ is in the risk set. The expectation is defined by $\mu_{ij}=\lambda_{ij}t_{ij} = \exp(g(\bfx_{ij},t_j) + \log(t_{ij}))$. For estimation, $\log(t_{ij})$ is included as an offset, thus the hazard rate $\frac{\mu_{ij}}{t_{ij}} = \lambda_{ij} = g(\bfx_{ij},t_j)$ is defined as the conditional expectation of having an event in interval $j$ divided by the time under risk. Note that the Poisson assumption is simply a computational vehicle for the estimation of the hazard \eqref{eq:pcmodel} rather than an assumption about the distribution of the event times. Despite the partition of the follow-up into intervals, this is a method for continuous event times as the information about the time under risk in each interval is contained in the offset and thus used during estimation. The number and placement of cut points controls the approximation of the hazard and could thus be viewed as a potential tuning parameter. In our experience, however, setting cut-points at the unique event times $\{t_i:\delta_i = 1,i=1,\ldots,n\}$ in the training data always leads to a good approximation (at least with enough regularization) as the number of cut-points will increase in areas with many events. For larger data sets, however, we recommend to set these cut-points on a smaller representative sub-sample of the data set (cf. Section~\ref{sec:compaspects}).

For the extension of \eqref{eq:phmodel} to MSMs, we define
\begin{equation}\label{eq:cshazard}
\lambda(t|\bfx, k) = \exp\left(f(\bfx(t), t,k)\right),\ k = 1,\ldots,K,
\end{equation}
as the transition specific hazard for the transition indexed by $k$, i.e., $k$ is an index of transitions  $m_k\ra m_k'$ where $m_k$ is the initial state and $m_k'$ a transient or absorbing state. The set of possible transitions is given by $\{m_k\ra m'_k: k = 1,\ldots, K\} \subseteq \{m\ra m': m,m'\in \{0,\ldots,M\}, m\neq m'\}$, where $M+1$ denotes the total number of possible states. $f(\bfx(t), t, k)$ is a function of potentially time-varying features $\bfx(t)$, including multivariate and/or non-linear effects. The dependency of $f(\bfx(t), t, k)$ on time $t$ (TVE) and transition $k$ (MSM) is expressed in terms of interactions by defining $\tilde{\bfx} := (\bfx(t), t, k)$ and $f(\bfx(t), t, k) = f(\tilde{\bfx}(t))$. Let $t_{i,k}$ be the event or censoring time w.r.t. transition $m_k\ra m'_k$ and $\delta_{i,k} \in \{0,1\}$ the respective transition indicator.
As extension of \eqref{eq:ped_status} we define
\begin{align}
\delta_{ij,k} & =
    \begin{cases}
        1 & t_{i,k} \in (\kappa_{j-1}, \kappa_j] \wedge \delta_{i,k} = 1\\\nonumber
        0 & \text{else,}
    \end{cases},
t_{ij,k} & =
    \begin{cases}
        t_{i,k}-\kappa_{j-1} & \delta_{ij,k}=1\\
        \kappa_{j}-\kappa_{j-1}& \text{else}\end{cases}.\nonumber
\end{align}
Table \ref{tab:data-trafo} shows how the data must be transformed in order to estimate \eqref{eq:cshazard} via PEMs for the competing risks setting, i.e., $k=1,\ldots,K$ is an index of transitions $m_k=0\ra m_k', m_k'\in \{1,\ldots,M\}$; a concrete example is given in Table \ref{tab:exampleData}. For each $i=1,\ldots,n$, there is one row for each interval the observation unit was under risk for a specific transition. Thus, one data set is created for each transition such that transitions to state $m_k'$ are encoded as $1$ and everything else, i.e., censoring and transition to other states is encoded as $0$. These transition-specific data sets,  each containing a feature vector with the transition index $k$, are then concatenated. Note that we used the same interval split points $\kappa_j$ for all transitions in Table \ref{tab:data-trafo}. However, it would also be possible to choose transition specific cut-points $\kappa_{j,k}$, or, more generally, even use multiple time-scales \cite{iacobelli_multiple_2013}. In the general multi-state setting, the number of observation units under risk might depend on the transition and the intervals visited by $i$ are defined by $(t_{i,m},\kappa_{j_{i,k}}], \ldots, (\kappa_{J_{i,k}-1},\kappa_{J_{i,K}}]$, where $t_{i,m}$ is the time-point at which $i$ enters state $m$.
\begin{table}[!ht]
\begin{center}
\caption{Data structure after transformation to the piece-wise exponential data format in the competing risks setting. Horizontal lines indicate a new observation unit $i=1,\ldots,n$. Double horizontal lines indicate a new transition indexed $k=1,\ldots,K$. As before, $J_i$ is the interval in which $i$ was observed last, i.e., $t_i \in (\kappa_{J_i-1}, \kappa_{J_i}]$. Features $x$ depend on time via $x_{i,p}(t) = x_{ij,p}\ \forall t\in (\kappa_{j-1},\kappa_j], p=1,\ldots,P$. This is not a strict requirement, as additional split points could be set at each time point a feature value is updated. See Table \ref{tab:exampleData} for a concrete example.}
\begin{tabular}{rr|rrrrrrr}
$i$ & $j$ & $\delta_{ij,k}$ &$t_j$ & $t_{ij}$ & $k$ & $x_{ij,1}$ & $\ldots$ & $x_{ij,P}$\\
\hline
$1$ & $1$ & $\delta_{11,1}$ & $t_1$ & $t_{11}$ & $1$ & $x_{11,1}$ & $\ldots$ & $x_{11,P}$\\
$1$ & $2$ & $\delta_{12,1}$ & $t_2$ & $t_{12}$ & $1$ & $x_{12,1}$ & $\ldots$ & $x_{12,P}$\\
$\vdots$ & $\vdots$ & $\vdots$ & $\vdots$ & $\vdots$ & $\vdots$ & $\vdots$ & $\vdots$ & $\vdots$\\
$1$ & $J_1$ & $\delta_{1J_1,1}$ & $t_{J_1}$ & $t_{1J_1}$ & $1$ & $x_{1J_1,1}$ & $\ldots$ & $x_{1J_1,P}$\\
\hline
$2$ & $1$ & $\delta_{21,1}$ & $t_1$ & $t_{21}$ & $1$ & $x_{21,1}$ & $\ldots$ & $x_{21,P}$\\
$\vdots$ & $\vdots$ & $\vdots$ & $\vdots$ & $\vdots$ & $\vdots$ & $\vdots$ & $\vdots$ & $\vdots$\\
\hline
$n$ & $1$ & $\delta_{n1,1}$ & $t_1$ & $t_{n1}$ & $1$ & $x_{n1,1}$ & $\ldots$ & $x_{n1,P}$\\
$\vdots$ & $\vdots$ & $\vdots$ & $\vdots$ & $\vdots$ & $\vdots$ & $\vdots$ & $\vdots$ & $\vdots$\\
$n$ & $J_n$ & $\delta_{nJ_n,1}$ & $t_{J_n}$ & $t_{nJ_n}$ & $1$ & $x_{nJ_n,1}$ & $\ldots$ & $x_{nJ_n,P}$\\
\hline
\hline
$1$ & $1$ & $\delta_{11,2}$ & $t_1$ & $t_{11}$ & $2$ & $x_{11,1}$ & $\ldots$ & $x_{11,P}$\\
$\vdots$ & $\vdots$ & $\vdots$ & $\vdots$ & $\vdots$ & $\vdots$ & $\vdots$ & $\vdots$ & $\vdots$\\
\hline
\hline
$\vdots$ & $\vdots$ & $\vdots$ & $\vdots$ & $\vdots$ & $\vdots$ & $\vdots$ & $\vdots$ & $\vdots$\\
\hline
\hline
$1$ & $1$ & $\delta_{11,K}$ & $t_1$ & $t_{11}$ & $K$ & $x_{11,1}$ & $\ldots$ & $x_{11,P}$\\
$\vdots$ & $\vdots$ & $\vdots$ & $\vdots$ & $\vdots$ & $\vdots$ & $\vdots$ & $\vdots$ & $\vdots$\\
\hline
\end{tabular}
\label{tab:data-trafo}
\end{center}
\end{table}

\begin{table}[!ht]
\caption{Data transformation for a hypotethical competing risks example  ($K=2$) for 3 subjects, $i=1,\ldots,3$. Subject
$i=1$ experienced an event of type $k=2$ at $t_1=1.3$, subject $i=3$ experienced an event of type $k=1$ at time $t_3=2.7$, subject $2$ was censored at $t_2=0.5$. Tables present the transformed data with intervals $(0,1], (1,1.5], (1.5, 3]$ for causes $k=1$ (left) and $k=2=K$ (right). These can be used to estimate cause specific hazards, by applying the algorithm to each of the tables separately or cause specific hazards with potentially shared effects, by stacking the tables and using $k$ as a feature.}
\begin{center}
\setlength{\tabcolsep}{0.5em}

\begin{tabular}{rr|rrrr}
$i$ & $j$ & $\delta_{ij}$ & $t_j$ & $t_{ij}$ & $k$\\
    \hline
1   & 1   & 0             & $1$   & 1        & 1  \\
1   & 2   & 0             & $1.5$ & 0.3      & 1 \\
\hline
2   & 1   & 0             & $1$   & 0.5      & 1  \\
\hline
3   & 1   & 0             & $1$   & 1        & 1  \\
3   & 2   & 0             & $1.5$ & 0.5      & 1  \\
3   & 3   & 1             & $3$   & 1.2      & 1  \\
\hline
\end{tabular}\hspace{5em}
\begin{tabular}{rr|rrrr}
$i$ & $j$ & $\delta_{ij}$ & $t_j$ & $t_{ij}$ & $k$\\
\hline
1   & 1   & 0             & $1$   & 1        & 2  \\
1   & 2   & 1             & $1.5$ & 0.3      & 2 \\
\hline
2   & 1   & 0             & $1$   & 0.5      & 2  \\
\hline
3   & 1   & 0             & $1$   & 1        & 2  \\
3   & 2   & 0             & $1.5$ & 0.5      & 2  \\
3   & 3   & 0             & $3$   & 1.2      & 2  \\
\hline
\end{tabular}
\label{tab:exampleData}
\end{center}
\end{table}

In Section \ref{sec:experiments} we evaluate the suggested approach using an implementation based on GBT that we refer to as GBT (PEM). As a concrete computing engine we used the extreme gradient boosting (XGBoost) library \cite{chen_xgboost:_2016} without any alterations to the algorithm. Therefore all features of the library can be used directly when estimating the hazard on the transformed data set. Note, however, that depending on the algorithm used, one must be able to specify an offset during estimation and potentially some other, algorithm or implementation specific settings. For example, when using XGBoost to estimate the GBT (PEM), the objective function needs to be set to the Poisson objective and the base score must be set to 1, because the default of 0.5 would imply a wrong offset, while $\log(1)=0$. The offset ($\log(t_{ij})$) must be attached to the data via the base margin argument during estimation. In contrast, for the prediction of the conditional hazard $\lambda(t|\bfx_i)=\lambda_{ij}$ based on new data points, the offset should be omitted, otherwise the algorithm will predict $\hat{\mu}_{ij}=\hat{\lambda}_{ij}\cdot t_{ij}$ (the expectation) instead of $\hat{\lambda}_{j}(\bfx)$ (the hazard). When predicting the cumulative hazard or survival probability, however, the time under risk in each interval must be taken into account, such that $\hat{S}(t|\bfx_i) =  \exp\left(-\int \hat{\lambda}(t|\bfx_i)\mathrm{d}t\right) = \exp\left(-\sum_{j=1}^{j(t)}\hat{\lambda}_{ij}\tilde{t}_{j}\right)$, where $j(t)$ indicates the interval for which $t \in (\kappa_{j-1}, \kappa_j]$ and $\tilde{t}_j=\min(\kappa_j-\kappa_{j-1}, t-\kappa_{j-1})$ is the time spent in interval $j$.
A prototype implementation of the GBT (PEM) algorithm that takes these issues into account and also provides the necessary helper functions for data transformation, estimation, tuning, prediction and evaluation is provided at \url{https://github.com/adibender/pem.xgb}.

\section{Experiments}\label{sec:experiments}
We perform a set of benchmark experiments with real world and synthetic data sets, exclusively using openly and directly available data, including a subset of data sets from recent publications on oblique random survival forests (ORSF, \cite{jaeger_oblique_2019}) and DeepHit \cite{lee_deephit_2018}. DeepHit and ORSF both have been shown to outperform other approaches such as RSF \cite{ishwaran_random_2008}, conditional forests \cite{hothorn_unbiased_2006}, regularized Cox regression \cite{friedman_regularization_2010} and DeepSurv \cite{ranganath_deep_2016}. We compare our approach in benchmarks against the two algorithms, which are evaluated separately based on evaluation measures used in the respective publications to ensure comparability.
All code to perform respective analyses as well as additional supplementary files are provided in a GitHub
repository: \url{https://github.com/adibender/machine-learning-for-survival-ecml2020}.\\

The data sets used for single event comparisons are listed in Table \ref{tab:single-event-data}.
The ``synthetic (TVE)'' data set is created based on an additive predictor $g(\bfx,t) = f_0(x_0, t)\cdot 6 -0.1\cdot x_1 + f_2(x_2, t) + f_3(x_3, t)$, where $f_0$, $f_2$ and $f_3$ are bivariate, non-linear functions of the inputs (see code repository for details) and $x_0, \ldots, x_3$ feature columns comprised in $\bfx$. Additionally, 20 noise variables are drawn from the uniform distribution $U(0, 1)$.

For the comparison with ORSF, we use the Integrated Brier Score $\text{IBS}(\tau) = \frac{1}{\tau} \int_{0}^\tau \widehat{\text{BS}}(u, \hat{S})\mathrm{d}u$, where $\widehat{\text{BS}}(t, \hat{S})$ is the estimated Brier Score at time $t$ weighted by the inverse probability of censoring weights \cite{gerds_consistent_2006} and $\hat{S}$ the estimated survival probability function of the respective algorithm. In addition, \cite{jaeger_oblique_2019} report the time-dependent C-Index \cite{gerds_estimating_2013}. We only consider the IBS here as it measures calibration as well as discrimination, while the C-index only measures the latter. Note that the IBS depends on the specific evaluation time $\tau$ and different methods might perform better at different evaluation times. Therefore, we calculate the IBS for three different time-points, the 25\%, 50\% and 75\% quantiles of the event times in the test data, in the following referred to as Q25, Q50 and Q75, respectively.

\begin{table}[ht]
\caption{Data sets used in benchmark experiments for comparison with ORSF.}
\centering
\begin{tabular}{rlrrr}
  \hline
 & name & n & p & censoring \\
  \hline
  1 & PBC & 412 & 14 & 61.90 \\
  2 & Breast & 614 & 1690 & 78.20 \\
  3 & GBSG 2 & 686 & 8 & 56.40 \\
  4 & Tumor & 776 & 7 & 51.70 \\
  5 & synthetic (TVE) & 1000 & 24 & $\sim 33\%$\\
   \hline
\end{tabular}
\label{tab:single-event-data}
\end{table}
For comparison with DeepHit, we use the metabric data set (cf. \cite{lee_deephit_2018}) for single event comparison as well as two CR data sets. The ``MGUS 2'' data is described in \cite{kyle_long-term_2002}. The ``synthetic (TVE CR)'' data set is simulated using an additive predictor identical to the one used for the ``synthetic (TVE)'' data simulation for the first cause. The predictor for the second cause, has a simpler structure $f_0(x_0, t) + 2\cdot x_4 -.1 \cdot x_5$, however, with non-proportional baseline hazard with respect to $x_0\in\{-1,1\}$. The number of noise variables is limited to 10 for this setting. Here we report the weighted C-index alongside  the weighted Brier Score as it was the main measure reported in \cite{lee_deephit_2018}. The proposed GBT (PEM) approach for CR (cf. Section \ref{sec:methods}) is a cause specific hazards model, however, the parameters of both causes are estimated jointly and the hazards of both causes can have shared effects (see Figure \ref{fig:i-can-do-the-splits-no-prob}). The simulation setting, therefore, constitutes a difficult setup because there are no shared effects and optimization w.r.t. the first cause will favor parameters that allow flexible models while the optimization w.r.t. the second cause favors sparse models and thus parameters that would restrict flexibility.


\begin{table}[ht]\label{tab:deephit-comparison-data}
\caption{Data sets used for the comparison with DeepHit. MGUS2 and Synthetic (TVE CR) are data set with two competing risks and additional right-censoring.}
\centering
\begin{tabular}{rlrrr}
  \hline
  & name & n & p & censoring (\%) \\
  \hline
  & METABRIC & 1981 & 79 & 55.20 \\
  & MGUS 2 & 1384 & 6 & 29.6\\
  & Synthetic (TVE CR) & 500 & 14 & $\sim 23\%$\\
  \hline
\end{tabular}
\end{table}

\subsection{Evaluation}
We compare four algorithms, the non-parametric Kaplan Meier estimate (Reference) as a minimal baseline, the Cox proportional hazards model \cite{cox_regression_1972} (baseline for linear, time-constant effects), the Oblique Random Survival Forest (ORSF) \cite{jaeger_oblique_2019} and {DeepHit} \cite{lee_deephit_2018}. For each experimental replication for a specific data set, 70\% of the data is randomly assigned as training data and the remaining 30\% is used to calculate the evaluation measures at three time points Q25, Q50 and Q75.  Algorithms are tuned on the training data using random search with a fixed budget and 4-fold cross-validation. Each algorithm is then retrained on the entire training data set using the best set of parameters before making final predictions on the test set.
The random search consists of 20 iteration for each algorithm. For the GBT (PEM), we define the search space  as follows (possible range in brackets): maximum tree depth $\{1,\ldots,20\}$, minimum loss reduction [0, 5], minimum child weight $\{5, \ldots, 50\}$, subsample percentage (rows) [0.5, 1], subsample percentage of features in each tree [.5, 1], L2-regularization [1, 3]. The learning rate is set to 0.05 and number of rounds to 5000, with early stopping after 50 rounds without improvement. For the ORSF we tune the elastic net mixing parameter (0, 1),
the parameter that penalizes complexity of the linear predictor in each node (0.25, 0.75), minimum number of events to split node $\{5,\ldots, 20\}$ and minimum observations to split node $\{10, 40\}$. For DeepHit, we use 50 random search iterations, where we search through \{1,2,3,5\} shared layers with \{50, 100, 200, 300\} dimensions, \{1,2,3,5\} cause-specific network layers with \{50,100,200,300\} dimensions, ReLU, eLU or Tanh as activation function in these layers, a batch size in \{32,64,128\}, a maximum of 50000 iterations, a dropout rate of 0.6 (taken from the original paper) and a learning rate of 0.0001. The network specific parameters $\alpha$ and $\gamma$ are also chosen in accordance with the original paper and set to $1$ and $0$, respectively, while the network specific parameter $\beta$ is varied in the random search with possible values in  \{0.1, 0.5, 1, 3, 5\}.


\subsection{Results}
The results for the experiments based on single-event scenarios comparison with ORSF
are summarized in Table \ref{tab:res-orsf}. The proposed method performs
well in many settings in comparison to ORSF. Notably, both algorithms are often not much better
than the Cox PH models indicating that the PH assumption is not violated strongly in those data sets and the sample size might be too small to detect small deviations
w.r.t. to non-linearity of feature effects, interaction effects and time-varying
effects. The ``synthetic (TVE)'' setting illustrates that in the presence of
strong, non-linear and non-linearly TVE our approach clearly outperforms the other methods.
For the PBC data we additionally ran an analysis including TVF with GBT (PEM). In this case, the inclusion of TVF resulted in a worse performance (IBS of 4.3 (Q25), 6.4 (Q50) and 9.2 (Q75)), which indicates that the inclusion of TVF lead to overfitting or that simple inclusion of the last observed value and carrying the last value forward is not appropriate in this setting.\\

\begin{table}
    \centering
\caption{Results of benchmark experiments for single event data comparing GBT (PEM) with ORSF. Bold numbers indicate the best performance for each setting.}
\begin{tabular}{llcccc}
\hline
data &  & Kaplan-Meier & Cox-PH & ORSF & GBT (PEM)\\
\hline
& Q25  & $\phantom{0}\textbf{1.9}$&\phantom{0}- & $\phantom{0}2.0$ & $\phantom{0}2.0$\\
\nopagebreak Breast  & Q50  & $\phantom{0}4.1$& $\phantom{0}-$ &$\phantom{0}\textbf{4.0}$  & $\phantom{0}\textbf{4.0}$\\
 & Q75  & $\phantom{0}7.2$& $\phantom{0}-$ &$\phantom{0}\textbf{6.7}$ & $\phantom{0}\textbf{6.7}$\\
 \hline
& Q25  & $\phantom{0}3.1$ & $\phantom{0}3.1$ & $\phantom{0}\textbf{2.9}$ & $\phantom{0}3.0$\\
\nopagebreak GBSG 2 & Q50  & $\phantom{0}6.8$ & $\phantom{0}6.5$ & $\phantom{0}\textbf{6.2}$ & $\phantom{0}6.4$\\
 & Q75  & $12.5$ & $11.4$ & $\textbf{11.1}$ & $11.3$\\
\hline
& Q25  & $\phantom{0}5.4$ & $\phantom{0}\textbf{3.7}$ & $\phantom{0}4.0$ & $\phantom{0}3.8$\\
\nopagebreak PBC  & Q50  & $\phantom{0}9.1$ & $\phantom{0}\textbf{5.3}$ & $\phantom{0}6.1$ & $\phantom{0}5.5$\\
 & Q75  & $14.0$ & $\phantom{0}8.1$ & $\phantom{0}8.6$ & $\phantom{0}\textbf{7.8}$\\
\hline
& Q25  & $\phantom{0}9.8$ & $\phantom{0}7.3$ & $\phantom{0}7.0$ & $\phantom{0}\textbf{4.6}$\\
\nopagebreak synthetic (TVE)  & Q50  & $19.2$ & $10.3$ & $\phantom{0}9.9$ & $\phantom{0}\textbf{6.7}$\\
 & Q75  & $23.7$ & $11.1$ & $11.7$ & $\phantom{0}\textbf{8.6}$\\
\hline
& Q25  & $\phantom{0}6.7$ & $\phantom{0}6.0$ & $\phantom{0}\textbf{5.5}$ & $\phantom{0}5.8$\\
\nopagebreak Tumor  & Q50  & $12.3$ & $11.2$ & $\textbf{10.8}$ & $10.9$\\
 & Q75  & $17.6$ & $16.3$ & $\textbf{16.2}$ & $\textbf{16.2}$\\
\hline
\end{tabular}
\label{tab:res-orsf}
\end{table}

Table \ref{tab:res-cr} summarizes the results of comparisons with DeepHit. The GBT (PEM) again shows good overall performance. For the synthetic data set our method  clearly outperforms the other approaches because it is capable of estimating non-linearity as well as time-variation. On the MGUS 2 data set, DeepHit shows the best performance for cause 1, while GBT (PEM) outperforms the other approaches for cause 2. On the synthetic data, the cause-specific Cox-PH model shows good discrimination (C-Index) for the second cause, but is worse than GBT (PEM) and DeepHit w.r.t. to the Brier Score.

\begin{table}[!ht]
\begin{center}
\caption{Results of benchmark experiments comparing GBT (PEM) with DeepHit for single event and CR data.
Bold numbers indicate the best performance for each setting.}\label{tab:res-cr}
\begin{tabular}{lll|ccc|ccc}
\hline
& & & \multicolumn{3}{c}{cause 1} & \multicolumn{3}{|c}{cause 2} \\
data & index & method & Q25 & Q50 & Q75 & Q25 & Q50 & \multicolumn{1}{c}{Q75} \\
\hline
 & & Cox-PH  & $13.3$ & $22.1$ & $26.4$ & $\phantom{00}-$ & $\phantom{00}-$ & $\phantom{00}-$ \\
 & Brier Score & DeepHit  & $14.3$ & $23.5$ & $27.0$ & $\phantom{00}-$ & $\phantom{00}-$ & $\phantom{00}-$ \\
 &  & GBT (PEM)  & $\textbf{12.8}$ & $\textbf{21.3}$ & $\textbf{25.9}$ & $\phantom{00}-$ & $\phantom{00}-$ & $\phantom{00}-$ \\
 \cline{2-9}
 \nopagebreak METABRIC & & Cox-PH  & $63.7$ & $65.1$ & $64.7$ & $\phantom{00}-$ & $\phantom{00}-$ & $\phantom{00}-$ \\
 & C-Index  & DeepHit  & $68.6$ & $63.3$ & $54.9$ & $\phantom{00}-$ & $\phantom{00}-$ & $\phantom{00}-$ \\
 &  & GBT (PEM)  & $\textbf{71.9}$ & $\textbf{71.5}$ & $\textbf{67.7}$ & $\phantom{00}-$ & $\phantom{00}-$ & $\phantom{00}-$ \\
\hline
\hline
 & & Cox-PH (CS)  & $23.6$ & $43.7$ & $64.3$ & $13.4$ & $20.5$ & $22.3$ \\
 & Brier Score  & DeepHit  & $\textbf{22.8}$ & $\textbf{41.0}$ & $\textbf{57.8}$ & $14.9$ & $27.0$ & $41.5$ \\
 \nopagebreak MGUS 2 &  & GBT (PEM)  & $22.9$ & $41.6$ & $60.5$ & $\textbf{13.0}$ & $\textbf{20.1}$ & $\textbf{22.1}$ \\
\cline{2-9}
 & & Cox-PH (CS)  & $66.7$ & $\textbf{65.9}$ & $\textbf{62.4}$ & $68.8$ & $69.4$ & $70.1$ \\
 & C-Index  & DeepHit  & $59.6$ & $57.0$ & $52.3$ & $65.5$ & $67.2$ & $68.6$ \\
 &  & GBT (PEM)  & $\textbf{68.4}$ & $62.9$ & $60.5$ & $\textbf{72.6}$ & $\textbf{70.9}$ & $\textbf{70.8}$ \\
\hline
\hline
&  & Cox-PH  & $\phantom{0}9.4$ & $13.1$ & $25.1$ & $35.5$ & $44.3$ & $50.6$ \\
 & Brier Score & DeepHit  & $\phantom{0}9.5$ & $16.0$ & $28.9$ & $33.0$ & $38.8$ & $\textbf{41.0}$ \\
 &  & GBT (PEM)  & $\phantom{0}\textbf{7.2}$ & $\textbf{11.6}$ & $\textbf{20.6}$ & $\textbf{30.1}$ & $\textbf{38.0}$ & $43.6$ \\
 \cline{2-9}
\nopagebreak synthetic (TVE, CR)  & & Cox-PH  & $90.2$ & $89.5$ & $85.4$ & $\textbf{86.5}$ & $\textbf{83.9}$ & $\textbf{81.6}$ \\
 & C-Index & DeepHit  & $92.3$ & $90.8$ & $84.6$ & $82.0$ & $80.1$ & $79.8$ \\
 &  & GBT (PEM)  & $\textbf{93.9}$ & $\textbf{92.2}$ & $\textbf{87.5}$ & $80.9$ & $80.8$ & $81.0$ \\
\hline
\end{tabular}
\end{center}
\end{table}

\section{Algorithmic Details and Complexity Analysis} \label{sec:compaspects}

We now briefly describe algorithmic details and discuss the complexity of the resulting algorithms when using the proposed framework.

\begin{figure}
    \centering
    \includegraphics[width=1\textwidth]{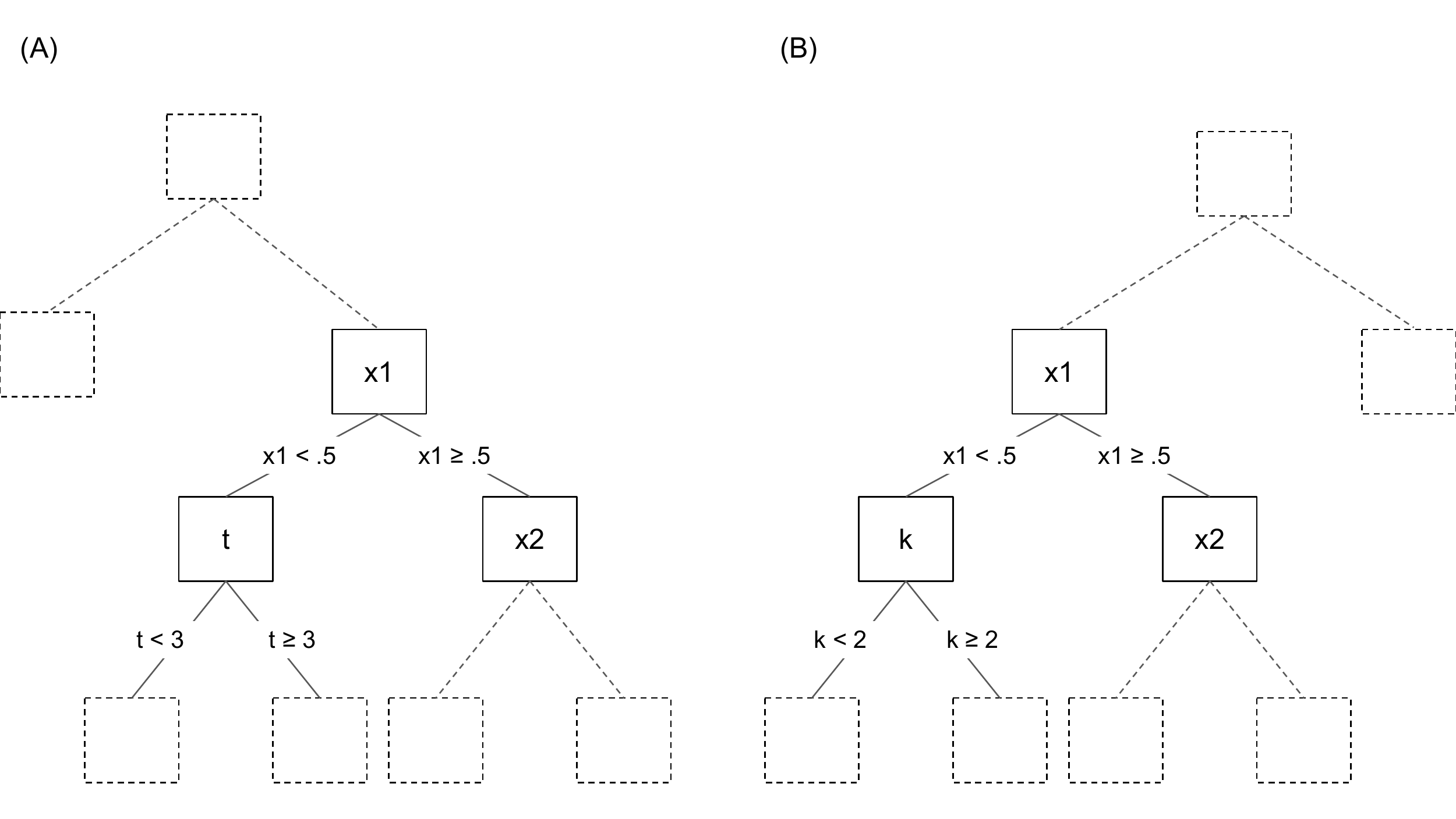}
    \caption{Illustration of how TVE and shared vs. transitions specific effects can be understood in terms of feature interactions in tree based models.}
    \label{fig:i-can-do-the-splits-no-prob}
\end{figure}

\paragraph{Algorithmic Details}
\ \\
The proposed framework is general in the sense that it transforms a survival task into a regression task. Nevertheless, different methods (and algorithms) have different strengths and weaknesses and different strategies can be applied to specify various alternative models within this framework. For example, in tree based methods, time-variation of feature effects could be controlled by allowing interactions of the time variable only with a subset of features, e.g. based on prior information, and similarly in order to control shared vs. transition specific effects in the multi-state setting. Tree-based methods are particularly intuitive when it comes to understanding the integration of TVE and extension to multi-state models via interaction terms into the model. This is illustrated in Figure \ref{fig:i-can-do-the-splits-no-prob}. For example, in panel (A) of Figure \ref{fig:i-can-do-the-splits-no-prob}, features and split points before the split w.r.t. time indicate feature effects common to all time-points. Once the data in panel (A) is split w.r.t. time $t$, the predicted hazard will be different for intervals with $\kappa_j < 3$ and $\kappa_j \geq 3$ for observations with $x_1 < .5$. Similarly, in a multi-state setting (panel (B) in Figure \ref{fig:i-can-do-the-splits-no-prob}), splits above the split w.r.t. $k$ indicate shared effects for all transitions, while splits below indicate different effects for transitions $k < 2$ vs. $k\geq 2$. Forcing a split w.r.t. to $k$ at the root node would be equivalent to an estimation of cause specific hazards on each subset and no shared effects.

Neural networks are particularly flexible when it comes to the specification of different PEMs. For example, the network could be split in two subnetworks, one for the temporal component, one for features, which is equivalent to the specification of a proportional hazards model, while allowing for non-linearity and high-dimensional interactions in feature effects. Similarly, defining subnetworks of the time variable for each category of a categorical feature would imply a stratified proportional hazards model.


\paragraph{Complexity Analysis}
\ \\
As described in the literature review, various approaches exist that account for special survival characteristics like TVF, CR or continuous time-scale prediction 
by altering the underlying method.
While adapting the structure of the algorithm itself potentially increases the complexity of the method,
our approach leaves the algorithm of choice unchanged as different time points and transitions are simply included as features. This allows to employ commonly used prediction methods without introducing further algorithmic complexity.
We note, however, that our approach might be improved upon in terms of scaling with respect to the number of intervals $J$ relative to the number of observations $n$. In the worst case, the number of total data points is quadratic in $n$ (or more precisely $\mathcal{O}(n(n+1)/2)$) when one interval cut-point is introduced for each observed event or censoring time.
We therefore propose a refinement of the presented method that improves run-times without forfeiting performance. Instead of setting cut-points at all unique event times, we suggest to define cut points more sparsely, for example, based on a sub-sample of the original data.

To investigate this strategy we conduct a scaling experiment where the sample size was consecutively doubled starting from $n=400$ up to $n=3200$. For each sample size, ten replications of one experiment as described for the ``synthetic TVE'' setting in Section \ref{sec:experiments} were performed and the elapsed time (hours) as well as performance (IBS) for two different strategies of cut-point selection was measured. The first strategy (full) uses all event times ($t_i$ where $\delta_i = 1$) as cut-points. The second strategy (sub-sample) is equivalent to the first strategy, but event times were chosen based on a sub-sample of $n'=200$, selected randomly from the training data in each iteration.
Results in Table \ref{tab:scaling-experiments} show that the ``sub-sample'' strategy leads to an approximately linear increase in computation time while the performance remains virtually unchanged. Potentially, a sparser choice of cut-points could also lead to a more robust and thus improved hazard estimation, as more events are available in each interval, but we did not conduct a formal investigation in that regard.

\begin{table}
\begin{center}
\caption{Results of the scaling experiments with $n$ the number of observation in the simulated data. ``strategy'' refers to the way interval cut-points were selected with ``full'' splits at all unique event times ($t_i$ where $\delta_i = 1$) and ``sub-sample'' refers to the selection of cut-points based on unique event times based on a random sub-sample of size $n'=200$ (but all observations were used for estimation). Mean time (in hours) and IBS over 10 replications are reported for each setting.}
\begin{tabular}{llcccc}
\hline
& & \multicolumn{4}{c}{n} \\
 & strategy & 400 & 800 & 1600 & \multicolumn{1}{c}{3200} \\
\hline
\nopagebreak time (hours) & \nopagebreak full  & $0.10$ & $0.48$ & $2.49$ & $8.94$ \\
 & \nopagebreak sub-sample  & $\textbf{0.09}$ & $\textbf{0.20}$ & $\textbf{0.51}$ & $\textbf{1.04}$ \\
\hline
\nopagebreak IBS & \nopagebreak full  & $8.10$ & $6.50$ & $6.40$ & $\textbf{5.90}$ \\
 & \nopagebreak sub-sample  & $\textbf{8.00}$ & $\textbf{6.40}$ & $\textbf{6.20}$ & $\textbf{5.90}$ \\
\hline
\end{tabular}
\label{tab:scaling-experiments}
\end{center}
\end{table}

\section{Conclusion}\label{sec:discussion}

We have presented a general machine learning framework for time-to-event analysis based on a data augmentation strategy that reduces a large variety of survival analysis tasks to the optimization of a Poisson likelihood. We demonstrated its versatility and state-of-the-art performance. The availability of Poisson regression for most machine learning frameworks provides additional practical advantages. For example, photon-ML \cite{zhang_photon_2016} is a scalable machine learning library for Apache Spark \cite{zaharia_spark} that has no native support for survival analysis, but implements generalized linear mixed models. Therefore, survival modeling with high cardinality random effects (frailty) is directly available using our framework. Similarly, lightGBM \cite{ke_lightgbm_2017}, a high-performance  implementation of GBT, currently has no implementation of survival methods, but could be also used for high-dimensional survival tasks based on PEMs, including reliability analysis or churn analysis with intermediate states.

\section*{Acknowledgements:}

This work has been funded by the German Federal Ministry of Education and Research (BMBF) under Grant No. 01IS18036A. The authors of this work take full responsibilities for its content.

%
%
\bibliographystyle{splncs04}
\bibliography{gdsurv.bib}

\end{document}